# Levels of AI Agents: from Rules to Large Language Models

Yu Huang
Roboaction.AI

**Abstract**:

*AI agents are defined as artificial entities to perceive the environment, make decisions and take actions. Inspired by the 6 levels of autonomous driving by SAE (Society of Automotive Engineers), the AI agents are also categorized based on utilities and strongness, as the following levels: **L0**—no AI, with tools (with perception) plus actions; **L1**—use rule-based AI; **L2**—let rule-based AI replaced by IL/RL-based AI, with additional reasoning & decision making; **L3**—apply LLM-based AI instead of IL/RL-based AI, additionally setting up memory & reflection; **L4**—based on L3, facilitating autonomous learning & generalization; **L5**—based on L4, appending personality (emotion + character) and collaborative behavior (multi-agents).*

## 1 Introduction

Any entity, that is able to perceive its environment and execute actions, can be regarded as an agent. Agents can be categorized into five types: **Simple Reflex agents, Model-based Reflex agents, Goal-based agents, Utility-based agents, and Learning agents** [1].

As AI advanced, the term "agent" is used to depict entities exhibiting intelligent behavior and possessing capabilities like autonomy, reactivity, pro-activeness, and social interactions. In the 1950s, Alan Turing proposed the renowned Turing Test [2]. It is a cornerstone in AI and aims to explore whether machines can show intelligent behavior comparable to human beings. These AI entities are usually called "agents", setting up the basic building blocks of AI systems.

Foundation models [3] have taken shape most strongly in NLP. On a technical level, foundation models are enabled by transfer learning and scale. The idea of transfer learning is to take the "knowledge" learned from one task and apply it to another task. Foundation models usually follow such a paradigm that a model is pre-trained on a surrogate task and then adapted to the downstream task of interest via fine-tuning.

Most of the Large Scale Language Models (LLMs) [4] appearing recently are among or based on the Foundation Models. Due to the remarkable capabilities exhibited recently, LLMs are considered as potential penetration of AI for Artificial General Intelligence (AGI), offering hope for building general AI agents.

An AI agent mostly refers to an artificial entity that is able to **perceive its surroundings using sensors, making decisions, and taking actions using actuators** [5]. According to the notion of World Scope (WS) [6] that audits the progress of NLP by encompassing 5 levels from NLP to general AI (i.e., Corpus, Internet, Perception, Embodiment, and Social), the pure LLMs-based agents are only built on the 2nd level from the written Internet world.

Except this, LLMs have proved remarkable capabilities in **knowledge capture, instruction interpretation, generalization, planning, and reasoning**, while showing natural language **interactions with humans**. From this status, the LLM assisted agents with an expanded perception space and action space, have the potential to reach the 3rd and the 4th levels of World Scope, i.e. Perception AI and Embodied AI respectively.

Moreover, these LLMs-based agents can handle more difficult tasks through **collaboration** or **gaming**, and social phenomena can be found, realizing the 5th level of World Scope, the Social World.

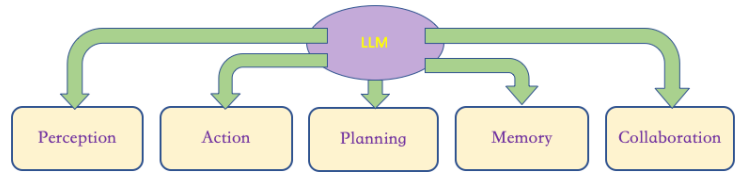

Fig. 1 LLM-based Agent

In session 2, LLMs is reviewed briefly; session 3 elaborates on various AI agents; levels of AI agents are analyzed and defined in session 4; and conclusion is given ae the end.

## 2 LLMs

LLMs [4] are the category of Transformer-based language models that are characterized by having an enormous number of parameters, typically numbering in the hundreds of billions or even more. These models are trained on massive text datasets, enabling them to understand natural language and perform a wide range of complex tasks, primarily through text generation and comprehension. Some well-known examples of LLMs include GPT-3/4, PaLM, OPT, and LLaMA1/2.

Extensive research has shown that scaling can largely improve the model capacity of LLMs. Thus, it is useful to establish a quantitative approach to characterizing the scaling effect. There are two representative scaling laws for Transformer language models: one from OpenAI[7], another from Google DeepMind[8].

The "pre-train+fine-tune" procedure is replaced by another procedure called "pre-train+prompt+predict" [9]. In this paradigm, instead of adapting pre-trained LMs to downstream tasks via objective engineering, downstream tasks are reformulated to look more like those solved during the original LM training with the help of a textual prompt.

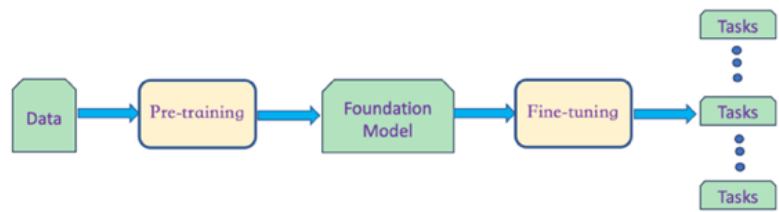

Fig. 2 The new paradigm as pre-training and fine-tuning in the foundation model

In this way, by selecting the appropriate prompts, the model behavior can be manipulated so that the pre-trained LM itself can be used to predict the desired output, sometimes even without any additional task-specific training. Prompt engineering [10] works by finding the most appropriate prompt to allow a LM to solve the task at hand.

The emergent abilities of LLMs are one of the most significant characteristics that distinguish them from smaller language models. Specifically, in-context learning (ICL)[11], instruction following[12] and reasoning with chain-of-thought (CoT)[13] are three typical emergent abilities for LLMs.

Parameter-efficient fine tuning (PEFT)[14] is a crucial technique used to adapt pre-trained language models (LLMs) to specialized downstream applications. PEFT can be divided into addition-based, selection/specification-based or reparameterization-based. It only needs fine-tuning a small subset of parameters, making it convenient for edge devices, and it can effectively mitigate the catastrophic forgetting problem.

Since LLMs are trained to capture the data characteristics of pre-training corpora (including both high-quality and low-quality data), they are likely to generate toxic, biased, or even harmful content for humans. It is necessary to align LLMs with human values, e.g., helpful, honest, and harmless. Reinforcement Learning from

Human Feedback (RLHF) [15] has emerged as a key strategy for fine-tuning LLM systems to align more closely with human preferences.

Motivated by the potential of LLMs, numerous multimodal LLMs (MLLMs) [16] have been proposed to expand the LLMs to the multimodal field, i.e., perceiving image/video input, and conversing with users in multiple rounds. Pre-trained on massive image/video-text pairs, the visual-language models can only handle image-level tasks, such as image captioning and question answering.

Building on the powerful pre-trained LLM weights, multimodal LLMs aim to handle multiple types of input beyond text. Multimodal LLMs have been widely applied to various tasks, such as image understanding, video understanding, medical diagnosis, and embodied AI etc.

It is regarded that LLMs are equipped with human-like intelligence and common sense to preserve the potential to make us closer to the Artificial General Intelligence (AGI) field. The emergence of LLMs is possibly the milestone of knowledge-driven agents, which perceive the environment and accumulate knowledge [17-27].

## 3 AI Agents

AI agents are capable to comprehend, predict, and response based on its training and input data. While these capabilities are developed and improved, it's important to understand their limitations and the effect of the underlying data they are trained on. There are some abilities of AI agent systems: 1) perceiving and predictive modeling. 2) planning and decision making. 3) self-learning and continuous improvement; 4) execution and interaction; 5) personal and collaborative.

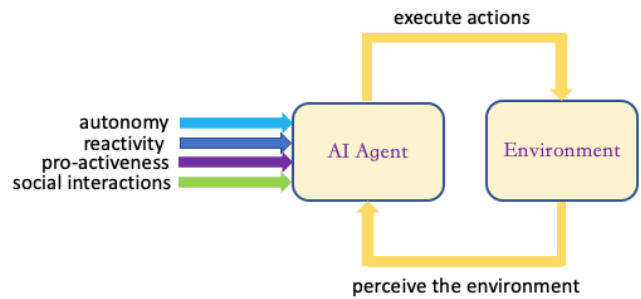

Fig. 3 Capabilities for an AI agent

The goal of embodied intelligence/AI is to build agents, such as robots, which learn to solve tasks with the need of interaction between agents and the environment.

For AI agents, an effective approach to learn actions, such as RL, is to carry out trial-and-error experiences via interactions with the environment. Training in the physical environment is often not feasible, so using simulators to learn policies is a common approach.

Symbolic AI [17-18] applied logical rules and symbolic representations to encapsulate knowledge and facilitate reasoning processes, in which critical problems are transduction and representation/reasoning. A classic example is knowledge-based expert systems. The symbolic agents confronted limitations in **uncertainty** and **large-scale** problems. They mainly relied on fixed algorithms or rule sets, working well in tasks they were built for. However, they were often difficult with **generalization** and **reasoning** when facing OOD (out-of-distribution) tasks.

RL-based agents [19-24] are trained by cumulative reward-based learning through interactions with their environments for handling more difficult tasks. An example is AlphaGo with **Q-learning**. Nonetheless, RL's problems are long training times, low sample efficiency, and stability concerns, especially for real-world environments.

Knowledge agents can utilize the knowledge implicitly or explicitly. Implicit knowledge is typically what LLMs encapsulate, and explicit knowledge is structured and queried to generate responses. The combination of implicit and explicit knowledge enable AI agents to apply knowledge contextually, akin to human intelligence.

LLM-based agents [25-35] employ LLMs as the primary component of brain or controller and expand their perceptual and action space through strategies such as multimodal perception and tool utilization. They can enable reasoning and planning abilities through techniques like **Chain-of-Thought (CoT)** and task splitting.

The LLMs has given rise to significant changes to AI agent design. These LLM agents are not only proficient in comprehending and producing natural language but also good at generalization. This ability enables easy integration with various tools, enhancing their versatility. On the other hand, the emergent abilities of LLMs shows the advantage at reasoning.

LLM agents, with pre-trained knowledge, have been prone to decision-making strategies even without task-specific training. On the other hand, RL agents often need to train from scratch in unseen cases, employing interaction to learn.

LLM-based agents can interact with each other, giving rise to the emergence of social phenomena. In LLM-based multi-agent systems (MAS), agents involve in collaboration, competition, or hierarchical platform to execute tasks. These tasks could start from search and optimization, decision making, and resource allocation to collaborative control.

The relations between agents govern the state of interaction and cooperation among them. Emotional reasoning and empathy are important skills for agents in many human-machine interactions.

## 4 Levels of AI Agents

Based on breath (generality) and depth (performance) of capabilities, a matrixed method to classifying AGI is given in [28] as below Table 1.

Table 1. Levels of AGI [28]

| Performance (rows) x Generality (columns) | Narrow clearly scoped task or set of tasks | General wide range of non-physical tasks, including metacognitive abilities like learning new skills |
|---|---|---|
| **Level 0: No AI** | **Narrow Non-AI** calculator software; compiler | **General Non-AI** human-in-the-loop computing, e.g., Amazon Mechanical Turk |
| **Level 1: Emerging** equal to or somewhat better than an unskilled human | **Emerging Narrow AI** GOFAI; simple rule-based systems, e.g., SHRDLU | **Emerging AGI** ChatGPT, Bard, Llama 2, Gemini |
| **Level 2: Competent** at least 50th percentile of skilled adults | **Competent Narrow AI** toxicity detectors such as Jigsaw; Smart Speakers such as Siri, Alexa, or Google Assistant; VQA systems such as PaLI; Watson; SOTA LLMs for a subset of tasks (e.g., short essay writing, simple coding) | **Competent AGI** not yet achieved |
| **Level 3: Expert** at least 90th percentile of skilled adults | **Expert Narrow AI** spelling & grammar checkers such as Grammarly; generative image models such as Imagen or Dall-E 2 | **Expert AGI** not yet achieved |
| **Level 4: Virtuoso** at least 99th percentile of skilled adults | **Virtuoso Narrow AI** Deep Blue, AlphaGo | **Virtuoso AGI** not yet achieved |
| **Level 5: Superhuman** outperforms 100% of humans | **Superhuman Narrow AI** AlphaFold, AlphaZero, StockFish | **Artificial Superintelligence (ASI)** not yet achieved |

**Performance** estimates how AGI compares to human-level performance for a given task. **Generality** measures the range of tasks for which an AI gets to a goal performance threshold. The rate of progression between levels of performance and/or generality may be nonlinear.

The desired features of Personal LLM Agents [35] need different kinds of capabilities. Inspired by the six levels of autonomous driving given by SAE（Society of Automotive Engineers）, the intelligence levels of Personal LLM Agents are categorized into 5 levels, from L1 to L5. The key characteristics and representative use cases of each level are listed in the Table 2 as below.

Table 2. Levels of Personal LLM Agents [35]

| Level | Key Characteristics | Representative Use Cases |
|---|---|---|
| L1 - Simple Step Following | Agent completes tasks by following exact steps pre-defined by the users or the developers. | - User: "Open Messenger";<br>- User: "Open the first unread email in my mailbox and read its content";<br>- User: "Call Alice". |
| L2 - Deterministic Task Automation | Based on the user's description of a deterministic task, agent auto-completes the steps in a predefined action space. | - User: "Check the weather in Beijing today". |
| L3 - Stratigic task Automation | Based on user-specified tasks, agents autonomously plan the execution steps using various resources and tools, and iterates the plan based on intermediate feedback until completion. | - User: "Make a video call to Alice". |
| L4 - Memory and Context Awareness | Agent senses user context, understands user memory, and proactively provides personalized services at times. | - User: "Tell the robot vacuum to clean the room tonight";<br>- User: "Tell Alice about my schedule for tomorrow". |
| L5 - Digital Persona | Agent represents the user in completing affairs, interacts on behalf of user with others, ensuring safety and reliability. | - User: "Find out which city is suitable for travel recently". |

In this paper, levels of AI Agents are defined based on utilities and strongness.

### 4.1 Tools (Perception + Action)

Various external tools support richer action capabilities for agents, including APIs, knowledge bases, visual encoding models and language models, enabling the agent to adapt to environmental changes, provide interaction and feedback, and even influence the environment. The tool execution can reflect the agents' complex requirements and augment the credibility of their decisions.

The action module targets at transforming the agent's decisions to specific outcomes. It interacts with the environment and gets feedback, deciding the agent's effectiveness in realizing tasks. The effectiveness of human feedback with the environment could enable agents to validate their action results.

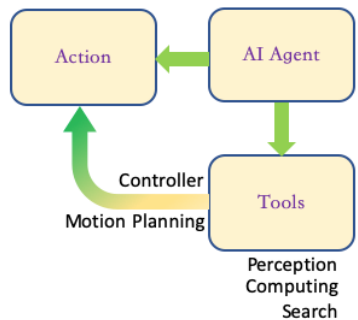

Fig. 4 Perception and Action for AI Agents

The action can have the perception module, the low-level motion planner and controller, especially in in robotics and autonomous driving. Especially, the perception module, like human sensory systems, i.e. the eyes and ears, perceives variations in the environment and then transfers multimodal information into an unified representation for the agent.

If the agent has been armed with the memory module, memory recollection can be an action strategy to enable agents making decisions based on experiences stored in memory modules.

The agent can leverage multiple rounds to determine appropriate responses as actions, especially for chat agents for the dialogue purpose.

The consequences of an action may encompass changes in the environment, changes in the internal agent states, kick-off of new actions, and influence on human perceptions in human-agent interaction scenarios.

## 4.2 Reasoning & Decision making

Reasoning is essential to human intelligence, working as the foundation for problem-solving, decision-making or planning, and critical analysis. Deductive, inductive, and abductive are the major forms of reasoning.

Conventional reasoning mainly relies on symbolic methods or imitation/reinforcement learning-based methods. But it is observed that those methods have several drawbacks. Symbolic methods need transferring natural language-described problems into rules, which may require manual help. Mostly, this type of method are sensitive to errors. Imitation Learning (IL) and Reinforcement Learning (RL) methods are usually combined with deep neural models, serving as the policy network, value function or reward model. While RL methods require huge samples (interactions with the environment), IL algorithms are hard to handle the unseen scenarios.

For LLM-based agents, like humans, reasoning is crucial for solving complex tasks. They may possess reasoning during pre-training or fine-tuning, meanwhile they may emerge after reaching a certain scale in size.

Task-decomposition is the most dominant form of planning in AI agents. It focuses on splitting a task into smaller sub-tasks and formulate specific plans for those. The key to solid task-decomposition techniques is to ensure there is a strong correlation between the created subtasks and the original tasks.

Ideally, subtasks should be independent of each other in task decomposition. That allows for the creation of prompts and contexts for each subtask, which makes troubleshooting easier by isolating failures to specific subtasks.

Chain-of-Thought (CoT) is the representative method of reasoning in LLMs, which solves complex reasoning problems step by step with a small number of language examples in the prompt. By decomposing complex tasks into executable sub-tasks, the ability of the LLMs to make plans and decisions is significantly improved.

Extensions of CoT include Tree-of-Thought (ToT) and Graph-of-Thought (GoT), assuming that humans tend to think in a tree-like or graph-like way. The multi-path thought further empowers the agent to solve more complex planning tasks.

The reasoning is undergone by planning or decision making. The planning module enables LLM-based agents with the ability to reason and plan for solving tasks, with or without feedback. Unlike conventional agents that call planning methods like Dijkstra, and POMDP to obtain the best actions and plan in the environments, RL-based agents need learning policies. LLM-based agents realize their planning capabilities from the LLM. Furthermore, LLMs display significant potential in intent understanding and other aspects.

LLM-based agents may fail to reach correct knowledge by prompts, or even face the hallucination problem. Specialized tools enable LLMs to boost their expertise and adapt domain knowledge. The decision-making process of LLM-based agents is short of transparency, less reliable in high-risk domains. Furthermore, LLMs are intolerant to adversarial attacks.

Tailoring the power of pre-trained models, with only a small amount of data for fine-tuning, LLMs can behave stronger performance in downstream tasks. Instead of only functioning as a fixed knowledge library, LLM-based agents demonstrate learning ability to adapt to new tasks robustly. Instruction-tuned LLMs show zero-shot generalization free from fine-tuning. LLMs can implement new tasks not occurred in the training stage, by following the instructions.

Few-shot in-context learning (ICL) upgrades the predictive performance of LLMs by combining the original input with few examples as prompts to enhance the context.

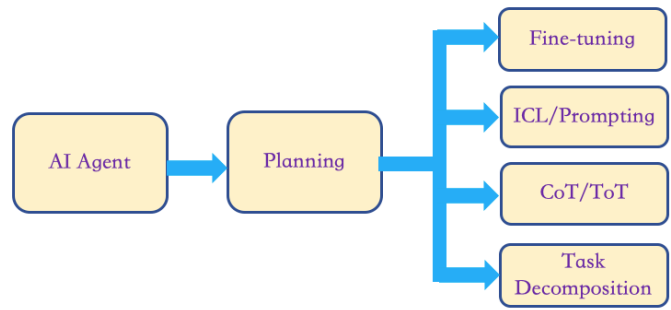

Fig. 5 Reasoning and Decision-making for AI agents

To emulate human capability on feedback experiences, planning modules can be designed to receive feedback from the environment, humans, and models, improving the LLMs-based agents' planning ability. External feedback serves as a direct evaluation of planning success or failure to build a closed-loop planning.

### 4.3 Memory + Reflection

The memory module takes a critical role in the AI agents. It stores information extracted from the environment perception and applies the stored memories to facilitate future actions. The memory module can help the agent to gather experiences, self-learn, and act in a more reasonable and effective manner.

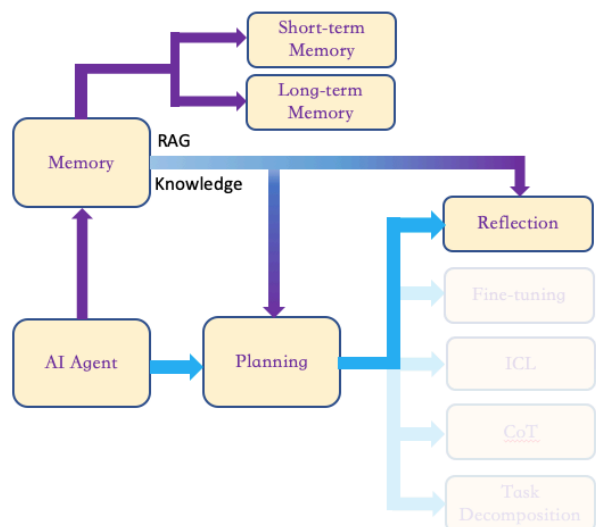

Fig. 6 Memory and Reflection for AI agents

Short-term memory keeps and preserves relevant information in the symbolic forms, guaranteeing its accessibility in the decision process. Long-term memory accumulates experiences from earlier decision process, consisting of history event flows, interaction information between the user and the agent or other formation of the agent's experiences.

The reflection module seeks to enable agents with the ability to compress and derive more advanced information, or verify and validate their actions autonomously. It helps agents in interpreting attributes, preferences, objectives, and connections, which in turn supervise their behaviors. It displays various forms: (1) self-summarization. (2) self-verification. (3) self-correction. (4) empathy.

Knowledge graph and retrieval augmented generation (RAG) could be additional modules for the memory mechanism. Knowledge graphs transforms data from multiple sources into a graph-based model of entities, their attributes, and how they relate. The graph can be continuously enriched by adding new data to uncover patterns and hidden relationships. To ensure AI agents that are powered by LLMs are responding with accurate, relevant responses, RAG is used to give LLMs domain-specific knowledge drawn from user manuals or support documents.

The agents assisted with LLMs, leverage internal feedback mechanisms, often giving rise to insights from pre-existing models, to polish and enhance planning approaches. They may obtain feedback from real or virtual surroundings, such as cues from task accomplishments or action responses, assisting them in revising and refining strategies.

## 4.4 Generalization & Autonomous Learning

Few-shot in-context learning (ICL) improves the LLMs' prediction by concatenating the original input with several examples as prompts to make the context stronger, which key idea is similar to the learning process of humans.

Instruction-tuned LLMs demonstrate zero-shot generalization with no need for fine-tuning on specific tasks. Prompting is critical for reasonable predictions, and training directly on the prompts can enhance the models' robustness for unseen tasks. The generalization level can be improved further by scaling up both the model size and the diversity of training instructions.

Agents need to generalize tool usage skills of users learned in contexts to new situations, such a model trained on Yahoo search as transferred to Google search.

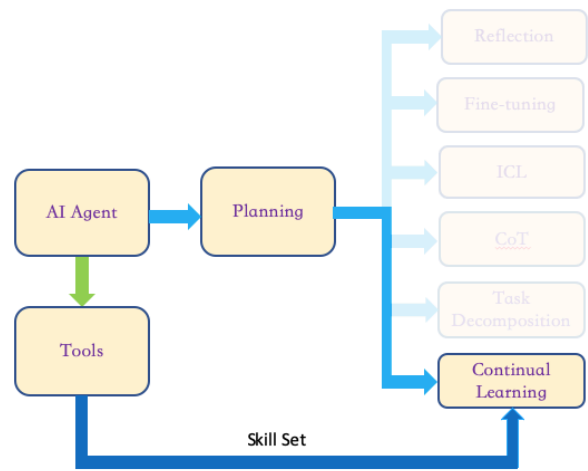

Fig. 7 Generalization and autonomous learning for AI agents

If instructions and demonstrations are given, LLM-based agents also hold the ability to build tools by generating executable programs, integrating current tools into stronger ones, or they can learn to perform self-debugging.

PaLM-E shows zero-shot or one-shot generalization capabilities to novel objects or combinations of existing objects. Voyager utilizes the skill library component to continuously gather new self-verified skills, which supports the AI agent's life-long learning capabilities.

LLM-based agents leverage the planning capabilities of LLMs to combine existing skills and handle more intricate challenges by continual learning, such as curriculum learning, to handle the challenges as catastrophic forgetting.

**4.5 Personality (Emotion + Character) and Collaborative Behavior (Multi-agents)**

Just as human personality happens through socialization, agents also demonstrate a form of personality through interactions with the others and the environment. The definition of personality refers to 3 features: cognitive, emotional, and character.

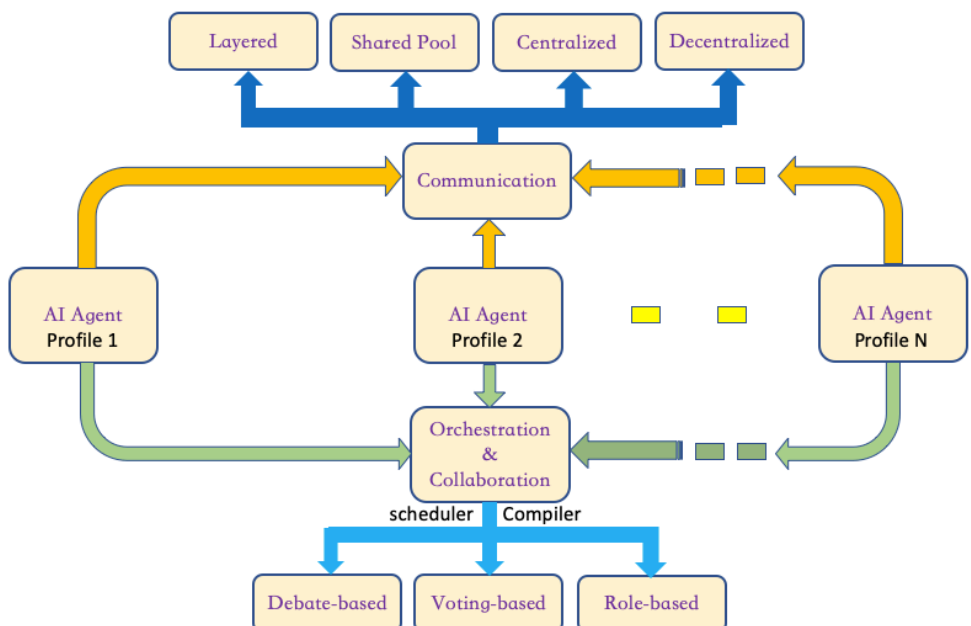

Fig. 8 Personality and collaboration for AI agents

Communication and collaboration are base stones for LLM-based Multi-AI agents. There are four typical communication structures in LLM-Multi-Agents systems. Layered communication is formulated hierarchically, with agents at each level playing different roles and interacting inter or intra layer. Decentralized communication operates on a peer-to-peer (P2P) network, where agents directly communicate with each other, a structure commonly employed in world simulation applications. Centralized communication involves a central agent or a group of central agents coordinating the system's communication, with other agents primarily interacting through this central node. Shared Message Pool could improve the communication efficiency by maintaining a shared message pool where agents publish messages and subscribe to relevant messages based on their profiles, thereby boosting communication efficiency.

The collaboration & orchestration methods could be classified as debate-based, voting-based and role-based. In the debate-based manner, each agent generates its own initial responses, and subsequently, a round of debate will start between the agents. In the voting-based way, an agent can first generate a candidate response to the user's prompts, then it holds a vote in which different reflective suggestions are presented as choices. In the role-based style, an agent-as-a-planner can generate a multi-step plan, an agent-as-an-assigner can orchestrate task assignment, while other tasks can be delegated to specific agent-as-a-worker based on their capabilities and expertise.

Cognitive abilities normally are defined as the mental processes in obtaining knowledge, such as decision making or planning, and problem-solving. Emotions include subjective moods, such as anger or joy. LLM-based agents include a detailed understanding of emotions.

The narrower concept of personality belongs to character patterns. Prompt engineering of LLMs involves the condensed summaries of character patterns or other attributes. Through exposure to personality enriched datasets, LLM-based agents are equipped with personality portrayal.

In social environments, AI agents should collaborate with or compete against other agents or even humans to provoke enhanced performance. The AI agents might be provided with complex tasks to co-work with or environments to interact with.

Collective intelligence is a process where the opinions are centralized into decisions. It comes from the collaboration and competition amongst agents, appearing in consensus-based decision-making patterns. By exploiting communication within an agent society, it becomes possible to emulate the evolution in human societies and gain insights.

### 4.6 Hierarchical Levels of AI Agents

Finally the levels of AI Agents are defined in Table 3.

Table 3. Levels of AI Agents

| AI Agent Levels | Techniques & Capabilities |
|---|---|
| L0: | No AI         + Tools (Perception + Actions) |
| L1: | Rule-based AI   + Tools (Perception + Actions) |
| L2: | IL/RL-based AI + Tools (Perception + Actions) + Reasoning & Decision Making |
| L3: | LLM-based AI  + Tools (Perception + Actions) + Reasoning & Decision Making + Memory & Reflection |
| L4: | LLM-based AI + Tools (Perception) + Actions + Reasoning & Decision Making + Memory & Reflection + Autonomous Learning + Generalization |
| L5: | LLM-based AI + Tools (Perception) + Actions + Reasoning & Decision Making + Memory+ Reflection + Autonomous Learning + Generalization + Personality (Emotion + Character) + Collaborative behavior (Multi-Agents) |

## 5 Conclusion

In this paper, levels of AI agents are categorized based on utilities and strongness, similar to automation levels of autonomous driving by SAE. For each level, the additional modules from the previous level could provide stronger AI capabilities and agent utilities. From level 0 to level 3, the AI core has evolved from no AI, to rule-based AI, IL/RL based AI to LLM-based AI.